\documentclass[lettersize,journal]{IEEEtran}
\usepackage{amsmath,amsfonts,amssymb}
\usepackage{algorithmic}
\usepackage{algorithm}
\usepackage{array}
\usepackage{textcomp}
\usepackage{stfloats}
\usepackage{url}
\usepackage{multirow}
\usepackage{bm}
\usepackage{verbatim}
\usepackage{needspace}
\usepackage{graphicx}
\usepackage{cite}
\usepackage{subfigure}
\usepackage{color}
\begin{document}

\title{AGI-Driven Generative Semantic Communications: Principles and Practices}
 
\author{Xiaojun Yuan, Haoming Ma, Yinuo Huang, Zhoufan Hua, Yong Zuo and Zhi Ding
\thanks{Xiaojun Yuan, Haoming Ma, Yinuo Huang and Zhoufan Hua are with the University of Electronic Science and Technology of China, China; Yong Zuo is with Xiangjiang Laboratory, China; Zhi Ding is with the University of California at Davis.}
}

\maketitle


\begin{abstract}

Semantic communications leverage artificial intelligence (AI) technologies to extract semantic information for efficient data delivery, thereby significantly reducing communication cost. With the evolution towards artificial general intelligence (AGI), the increasing demands for AGI services pose new challenges to semantic communications. In this context, an AGI application is typically defined on a general-sense task, covering a broad, even unforeseen, set of objectives, as well as driven by the need for a human-friendly interface in forms (e.g., videos, images, or text) easily understood by human users.In response, we introduce an AGI-driven communication paradigm for supporting AGI applications, called generative semantic communication (GSC). We first describe the basic concept of GSC and its difference from existing semantic communications, and then introduce a general framework of GSC based on advanced AI technologies including foundation models and generative models. Two case studies are presented to verify the advantages of GSC. Finally, open challenges and new research directions are discussed to stimulate this line of research and pave the way for practical applications.
\end{abstract}

\begin{IEEEkeywords}
Semantic communication, artificial intelligence, generative model, foundation model.
\end{IEEEkeywords}

\section{Introduction}
The rapid advancement of artificial intelligence (AI) has evidenced its immense potential in problem-solving, sparking significant interest in integrating AI technologies in various fields, especially in wireless communications. Early approaches involve applying AI to typical wireless problems, such as channel estimation, signal detection, and channel coding design \cite{zhu2020toward}. Recently, semantic communication has emerged as an intelligent communication scheme, prioritizing the meanings of data over traditional bit streams for transmission. For instance, in the task of language translation, a phrase like “the cat is on the mat” may be received as “the feline is on the rug” in semantic communication, as opposed to the original phrase “the cat is on the mat” received and reconstructed in a traditional communication system. The focus on the underlying meaning of data allows semantic communication to significantly reduce the overall communication cost \cite{dai2022communication}.

As AI continues to evolve, the increasing demand for artificial general intelligence (AGI) introduces new challenges and opportunities in the realm of semantic communications. Unlike early AI applications, AGI is typically defined on a general-sense task that involves a broader range of objectives, including even unforeseen objectives. Besides, an AGI application often requires a human-friendly interface in practical scenarios, ensuring that outputs of an intelligent system are easily understandable by humans, such as in the form of videos, images, and/or texts. For example, in telemedicine scenarios, doctors often prefer medical images annotated with information about the affected areas, rather than raw physiological data that is difficult to interpret, in order to make quick diagnoses. Similarly, in autonomous driving scenarios, human drivers often prefer semantic scene representations, highlighting lanes, pedestrians, and obstacles, instead of raw sensor data, thereby enabling them to intuitively monitor their surroundings. To support AGI services, it is essential to design efficient communication strategies for the fulfillment of miscellaneous task objectives, as well as to create human-friendly interfaces at user ends. However, existing semantic communications, typically defined on a predefined objective of a specific task, are difficult to meet the demands of AGI-driven communications, which calls for a more advanced semantic communication paradigm.


In response, we introduce AGI-driven generative semantic communication (GSC) as an example for supporting AGI applications.\footnote{The terminology GSC has been recently used in \cite{chang_gensc_2024, grassucci2023generative}, where generative models are employed at the receiver side for the \textit{exact reconstruction} of source data. In this regard, AGI-driven GSC introduced in this article is a more general concept that allows a receiver to generate perceptually plausible data that contain necessary semantic information of the source data, so as to substantially reduce communication overhead.} The key to meeting the requirements of an AGI application in GSC lies in the use of cutting-edge AI technologies such as foundation models and generative models. Specifically, a foundation model like DeepSeek and OpenAI GPT, known for its proficiency in understanding human needs, is utilized at the transmitter to comprehend and extract the information necessary to support diverse objectives of an AGI application. A generative model like diffusion model \cite{cao2024survey, chen_control--video_2023}, capable of synthesizing new data that resembles a natural sample from the perspective of human perception, is employed at the receiver to create a human-friendly interface. Interestingly, we observe that the information required for AGI applications exhibits different levels of distortion tolerance, enabling the compression of different types of information with different levels of information loss. In this article, we first introduce the basic concept of GSC. Then, we establish its general framework, elaborating on core components such as the semantic encoder and the semantic decoder, along with problem formulation. After that, we present two case studies that explore the specific implementation of GSC in an online meeting scenario and in a road monitoring scenario. Finally, we conclude the article and discuss several open challenges for GSC. 

\section{Fundamentals of Semantic Communications}\label{fundamental}

\subsection{Semantic Information}


\begin{figure*}
    \centering
    \includegraphics[width=0.9\linewidth]{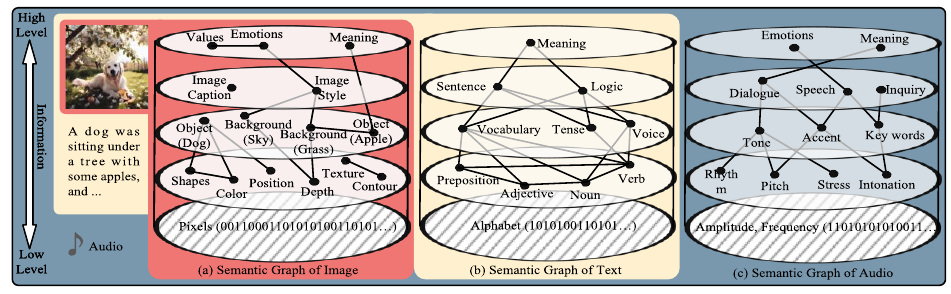}
    \caption{Illustration of semantic graphs corresponding to multi-modal data. (a) Semantic graph corresponding to an image of a dog sitting under a tree with several apples on the ground. (b) Semantic graph of text. (c) Semantic graph of audio.} In each graph, circles represent individual semantic nodes ($s \in \mathcal{S}$) and edges denote the relationships $(s, s^\prime) \in \mathcal{R}$ between these nodes.
    \label{semantic_hierarchy}
\end{figure*}

Semantic information in data typically exhibits an intrinsic structure known as a \textit{semantic graph}, where semantic attributes and their connections can be organized as a graph. Formally, a semantic graph is defined as $\mathcal{G}=(\mathcal{S},\mathcal{R})$, where $\mathcal{S}=\{s_1,\dots,s_N\}$ denotes the set of $N$ semantic nodes (also referred as semantics), and $\mathcal{R}=\{(s, s^\prime)\mid s, s^\prime\in\mathcal{S}, s\neq s^\prime\}$ denotes the set of edges representing relationships between these semantic nodes.
To illustrate the concept of a semantic graph, we present examples in Fig. \ref{semantic_hierarchy}. Particularly, Fig. \ref{semantic_hierarchy} (a) shows the semantic graph of an image of a dog sitting on the grass under a tree in a garden, where circles represent individual semantic nodes $s \in \mathcal{S}$ and edges represent the relationships $(s, s^\prime) \in \mathcal{R}$ between these nodes.
Considering that different semantic nodes exhibit varying levels of semantic granularity and abstraction, we empirically classify the inherent semantic nodes within the image into four hierarchical levels. For example, in Fig. \ref{semantic_hierarchy} (a), these levels range from low-level semantics, such as “shape” and “color coding” nodes, to higher-level semantics, including “value” and “emotion” nodes. Moreover, this semantic graph-based representation extends naturally to other types of data, like text and audio, as shown in (b) and (c) of Fig. \ref{semantic_hierarchy}, respectively. Particularly, in an audio-based speech comprehension task, high-level nodes such as semantic meaning and logical relationships are essential, whereas low-level nodes like exact word choice or pronunciation nuances are typically less critical. It is important to note that the presented semantic graphs may be far from complete; here, we present only a limited set of semantic nodes and relationships as representative instances. Such structural representation of information plays a pivotal role in understanding semantic communications, as detailed in what follows.


\subsection{Semantic Communication}


In the context of semantic graphs, traditional communications are primarily focused on transmitting the lowest level of information, such as pixels. For example, in image transmission, the pixel-level representation is sent for reconstruction, without requiring the transmitter or receiver to understand the image content.
While pixel-level transmission has been widely adopted in traditional communications, emerging intelligent tasks, such as object detection, rely more on higher-level information, implying the potential for performance improvement by bypassing the direct transmission of the lowest-level information. 
Semantic communications have been developed for this purpose \cite{zhu2020toward, dai2022communication}. For instance, in face recognition, a camera extracts facial features from an image and sends them to a server for face recognition. Since higher-level semantic information typically requires a lower information rate, semantic communication significantly reduces the communication overhead in intelligent applications, as compared to traditional communications.



A widely recognized semantic communication paradigm, known as task-oriented semantic communication (TOSC) \cite{kountouris2021semantics, Cai2024a, Huang2024}, has been proposed to facilitate a type of intelligent tasks referred to as “narrow-sense tasks”. 
In particular, a narrow-sense task is defined on a specific objective, and TOSC focuses on conveying only the information relevant to the objective, thereby reducing communication overhead. From the perspective of the semantic graph, the task-relevant information in TOSC is represented as an induced subgraph  $\mathcal{G}_\text{TOSC} \triangleq (\mathcal{S}_{\text{TOSC}}, \mathcal{R}_{\text{TOSC}})$, extracted from the full semantic graph $\mathcal{G}$. More formally, the set of semantic nodes relevant to a narrow-sense task is defined as $\mathcal{S}_{\text{TOSC}}=\{s\in \mathcal{S}| s \text{ is related to narrow-sense task } T\}$, ensuring that $\mathcal{S}_{\text{TOSC}}$ includes only those semantic nodes directly associated with task $T$.  The set of relations within the subgraph, $\mathcal{R}_{\text{TOSC}} = {(s, s^\prime) \mid s, s^\prime \in \mathcal{S}_{\text{TOSC}}, (s, s^\prime) \in \mathcal{R}}$, includes only the relations among the selected nodes that exist in $\mathcal{G}$. In practice, only the task-relevant information $\mathcal{G}_\text{TOSC}$ needs to be transmitted to the receiver to support the narrow-sense task. For example, in the case of an animal recognition task, where the objective $T$ is to identify and classify different animal species. The task-relevant information of TOSC $\mathcal{G}_\text{TOSC}$ includes nodes such as “object (dog)”, “shapes” and “color,” along with their relations inherited from the original graph. During transmission, semantic information like shape descriptors, color codes, and object coordinates is conveyed to the receiver.



Despite its advantages in communication efficiency, semantic communication faces new challenges as AI evolves towards AGI. An AGI application is typically defined on a general-sense task, which includes a wide range of objectives, including even unforeseen ones. This is in contrast to the narrow-sense task of TOSC that focuses on specific, pre-defined objectives. For example, in natural language processing, machine translation is a narrow-sense task, while comprehending and responding to natural language queries exemplifies a general-sense task, as it can accommodate diverse objectives based on user queries. Another key property of AGI applications is the frequent interaction of an AGI server with human users. This requires a human-friendly interface, meaning that the system’s outputs should align with human perception and remain readily understandable by humans, such as in the form of videos, images, and/or texts. However, simply transmitting semantic information for a particular narrow-sense task, as in TOSC, cannot meet the new demands of AGI applications. These challenges call for the development of a new semantic communication paradigm, as discussed below.

\subsection{AGI-Driven Generative Semantic Communication}

To meet the demands of AGI applications, we introduce a general semantic communication paradigm, named AGI-driven generative semantic communication (GSC), which supports a wide range of objectives inherent to a general-sense task, as in contrast to TOSC which caters to a narrow-sense task. From a semantic graph perspective, the task-relevant information for GSC is represented as an induced subgraph $\mathcal{G}_\text{GSC} \triangleq (\mathcal{S}_{\text{GSC}}, \mathcal{R}_{\text{GSC}})$, obtained from $\mathcal{G}$. Specifically, the set of semantic nodes relevant to a general-sense task is defined as $\mathcal{S}_{\text{GSC}}=\{s\in \mathcal{S}\mid s \text{ is related to a general-sense task }\mathcal{T}\}$, where the general-sense task $\mathcal{T}$ needs to be determined before transmissions and includes every potential task objective that an AGI application might require. For example, consider a telemedicine service, $\mathcal{T}$ includes various task objectives, such as “real-time monitoring of patient health status” and “providing a summary of patient medical history,” each of which corresponds to specific semantic nodes, such as “heart rate” and “blood pressure,” that together form the task-relevant information $\mathcal{G}_\text{GSC}$. Then, the set of relations within the induced subgraph is defined as $\mathcal{R}_{\text{GSC}}=\{(s, s^\prime)\mid s, s^\prime\in\mathcal{S}_{\text{GSC}},(s, s^\prime)\in\mathcal{R}\}$, containing only the relations between the selected nodes. Notably, a key difference between TOSC and GSC is that $\mathcal{G}_{\text{GSC}}$ is derived from a general-sense task $\mathcal{T}$, whereas $\mathcal{G}_{\text{TOSC}}$ is based on a certain narrow-sense task $T$.


Furthermore, to create a human-friendly interface, AGI-driven GSC is required to transmit perceptual information that, while not directly relevant to a general-sense task, is critical for producing outputs aligned with human perception.\footnote{In our proposed GSC framework, aligning with human perception essentially means that the distribution of the outputs at the receiver closely matches that of real-world (natural) data, i.e., data as perceived or interpreted by humans. Therefore, beyond creating a human-friendly interface, our proposed GSC framework can be applied generally to scenarios in which matching data distributions is critical. This issue will be elaborated later in Section \ref{performance}.} In the context of semantic graph, this perceptual information is represented as an additional subgraph $\mathcal{G}^{\star}_\text{GSC}\triangleq(\mathcal{S}_{\text{GSC}}^{\star}, \mathcal{R}_\text{GSC}^{\star})$, where the node set is defined as $\mathcal{S}_{\text{GSC}}^{\star}=\{s\in\mathcal{S}\mid s \text{ is related to human-perception}\}$, which includes all nodes that are necessary for human's perceptual understanding of natural data. Similarly, the relation set consists of all connections between these nodes, formally defined as $\mathcal{R}_{\text{GSC}}^{\star}=\{(s, s^\prime)\mid s, s^\prime\in\mathcal{S}_{\text{GSC}}^{\star},(s, s^\prime)\in \mathcal{R}\}$. By transmitting both the task-relevant information $\mathcal{G}_\text{GSC}$ and the perceptual information $\mathcal{G}^{\star}_\text{GSC}$, GSC enables the receiver to generate outputs that align with human perception. Since not all the elements in $\mathcal{G}^{\star}_\text{GSC}$ are equally useful, GSC may, in practice, transmit only a subset of this additional subgraph in a highly compressed form.


Overall, our AGI-driven GSC framework unifies existing semantic communication schemes by transmitting different types of semantic information with different levels of distortion tolerance. As an extreme example, GSC reduces to TOSC when it extracts task-relevant information $\mathcal{G}_\text{GSC}$ based on a narrow-sense task $T$ rather than a general-sense task $\mathcal{T}$, and at the same time omits all perceptual information $\mathcal{G}_\text{GSC}^\star$. On the other extreme, if all the information in source data is required at the receiver, our proposed AGI-driven GSC degenerates into semantic communication approaches for exact source data reconstruction, such as those in \cite{chang_gensc_2024, grassucci2023generative}. As such, AGI-driven GSC can flexibly balance the transmission of task-relevant and perceptual information, tailoring its output to diverse application needs.


\section{Framework of AGI-Driven Generative Semantic Communication}\label{framework}

\begin{figure*}
  \centering
  \includegraphics[width=0.95\linewidth]{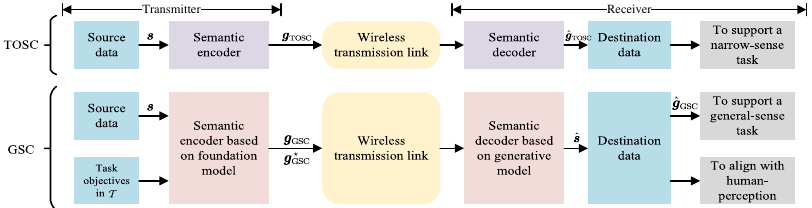}
  \caption{Flow diagrams illustrating the semantic communication frameworks for TOSC and GSC.}
  \label{semantic_framework}
\end{figure*}

\subsection{Overview}
In this section, we introduce the GSC framework designed to efficiently transmit both task-relevant and perceptual information. To achieve this, these two types of information are typically embedded into vector spaces rather than represented in form of a graph, enabling more compact and efficient transmission. As illustrated in Fig. \ref{semantic_framework}, the framework operates within a wireless communication scenario that involves a transmitter and a receiver. At the transmitter, a semantic encoder, under the guidance of a general-sense task $\mathcal{T}$, extracts both task-relevant information  $\boldsymbol{g}_\text{GSC}$ and perceptual information $\boldsymbol{g}^\star_\text{GSC}$ from the source data $\boldsymbol{s}$, embedding them in a vector space for transmission. Upon reception, the semantic decoder generates the destination data $\hat{\boldsymbol{s}}$ that is not designed for exact reconstruction of source data $\boldsymbol{s}$. Instead, $\hat{\boldsymbol{s}}$ is constrained to align with human perception and to fulfill the general-sense task $\mathcal{T}$, as required by AGI-driven applications. To fulfill these two requirements, we incorporate advanced AI technologies, including foundation models and generative models, to design the semantic encoder and the corresponding decoder in the GSC framework, as detailed in what follows.


\subsection{Semantic Encoder Based on Foundation Models}
In the GSC framework, the transmitter employs a semantic encoder to intelligently extract both task-relevant information $\boldsymbol{g}_\text{GSC}$ and perceptual information $\boldsymbol{g}^\star_\text{GSC}$ from the source data $\boldsymbol{s}$, based on the various objectives in $\mathcal{T}$, as illustrated in Fig. \ref{semantic_framework}.
This necessitates a versatile semantic encoder capable of extracting diverse information across different task objectives.
Foundation models, which are large-scale models pre-trained on massive and diverse datasets, are particularly well-suited for this role. This broad pre-training enables foundation models to learn general-purpose representations that can be adapted to new tasks with minimal or no additional training. As a result, a foundation model can flexibly serve as a universal semantic encoder across a wide range of objectives without retraining task-specific networks. For example, in an autonomous driving scenario, a foundation model can extract all the information required by the scenario, such as object categories and bounding-box coordinates, as the task-relevant information. In this case, the foundation model also extracts the depth maps of road images as perceptual information. Notably, emerging foundation models like GPT, LLaMa, and Palm-E have shown a strong ability to handle multimodal data including images, videos, audios, and texts \cite{yang_depth_2024, li_blip-2_2023}.




\subsection{Semantic Decoder Based on Generative Models}

As shown in Fig. \ref{semantic_framework}, the receiver of GSC uses a semantic decoder at the receiver to reconstruct the destination data $\hat{\boldsymbol{s}}$ that contains both task-relevant information and perceptual information. It is important to note that these two types of information represent only a subset of the semantic information from the source data, typically omitting a large amount of details to improve communication efficiency. However, these excluded details remain essential for creating a human-friendly interface. Instead of transmitting these details directly, the GSC framework proposes to employ a generative model at the receiver to generate the missing details. This generative model learns patterns from existing datasets and fills in the omitted details, ensuring that the final output aligns with human perception.\footnote{We clarify that foundation models and generative models are different concepts in machine learning. Of course, some models, such as generative pre-trained transformers (GPT), serve as both foundation models and generative models, since they are pretrained on large-scale datasets and can generate new content.} For instance, in an object recognition task, class labels and depth maps can serve as the task-relevant information and the perceptual information to be transmitted, respectively. In contrast, details such as color and texture need not to be transmitted, as they can be effectively generated at the receiver using a generative model. Generative models have evolved through three key stages \cite{cao2024survey}. Early attempts like variational auto-encoder (VAE) established basic generation through latent space encoding, yet suffered from limited output quality. Later, the generative adversarial network (GAN) introduced adversarial training between a generator and a discriminator, achieving improved image synthesis at the cost of training instability. More recently, diffusion-based models \cite{cao2024survey, chen_control--video_2023} have emerged as the state-of-the-art paradigm, demonstrating remarkable generation quality through innovative noise-adding and recursive denoising processes. This diffusion-based paradigm not only enables photorealistic image synthesis but also extends to video generation, establishing itself as the primary technique used in our proposed GSC framework.

\begin{figure*} 
  \centering
  \includegraphics[width=0.95\linewidth]{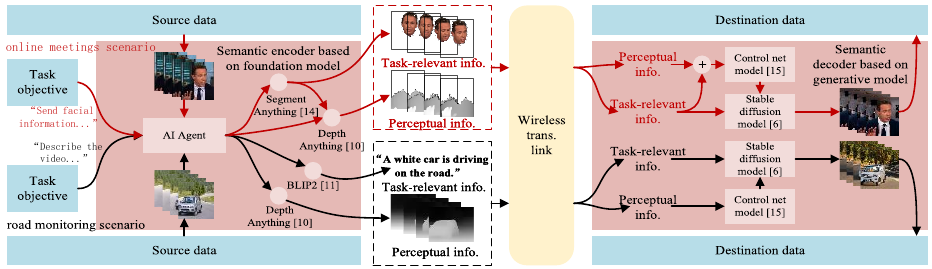}
  \caption{Flow diagrams of the GSC framework in two scenarios: online meetings and road monitoring.}
  \label{semantic_practical}
\end{figure*}

\subsection{Problem Formulation}\label{performance}
We are now ready to formulate the fundamental problem of AGI-driven GSC based on a modified rate-distortion-perception (RDP) theory. This problem quantifies trade-offs among communication rate, transmission distortion, and the perceptual quality of the reconstructed data. The problem objective is to minimize the overall transmission rate under two constraints, referred to as the task-relevant and perceptual constraints.\footnote{For simplicity of discussion, the formulation here does not account for physical-layer constraints, such as transmitter power and bandwidth limitations.} Specifically, the task-relevant constraint ensures that the distortion term $d(\hat{\boldsymbol{g}}_\text{GSC},\boldsymbol{g}_\text{GSC})$ stays below a predefined threshold, where $\hat{\boldsymbol{g}}_\text{GSC}$ denotes the task-relevant information extracted from the destination data $\hat{\boldsymbol{s}}$ at the receiver, and $\boldsymbol{g}_\text{GSC}$ denotes the task-relevant information extracted from the source data $\boldsymbol{s}$ at the transmitter. This constraint ensures that the transmission distortion of the task-relevant information remains sufficiently small, since a general-sense task is typically sensitive to such distortion. In practice, we may take a semantic-NMSE approach, which defines metric $d(\cdot,\cdot)$ as the normalized mean-square error (NMSE), and estimates $\hat{\boldsymbol{g}}_\text{GSC}\triangleq q(\hat{\boldsymbol{s}})$, where $q(\cdot)$ denotes the semantic encoder in the GSC framework. Importantly, compared with the classic RDP theory, which measures distortion by the distance $d(\boldsymbol{s},\hat{\boldsymbol{s}})$ between the source data $\boldsymbol{s}$ and its reconstruction $\hat{\boldsymbol{s}}$, our task-relevant constraint focuses only on the distortion of the task-relevant information.

Moreover, the perceptual constraint requires that the perceptual quality metric $p(\hat{\boldsymbol{s}})$ be less than a predefined threshold, where $p(\cdot)$ quantifies the perceptual quality of the destination data $\hat{\boldsymbol{s}}$.\footnote{The existing GSC schemes \cite{chang_gensc_2024, grassucci2023generative} focus on minimizing the reconstruction error between $\hat{\boldsymbol{s}}$ and the original source data $\boldsymbol{s}$. This is in contrast with our AGI-driven GSC.} In this way, the perceptual constraint aligns the system output $\hat{\bm{s}}$ with human perception. In practice, we can use various perceptual metrics to define $p(\cdot)$. For example, the no-reference quality metric (NRQM) \cite{ma2017learning} and the perception-based image quality evaluator (PIQE) \cite{assesment2022} are widely used image-quality metrics. 

It is noteworthy that our GSC framework supports use cases far beyond creating a human-friendly interface. For example, in the case of transmitting a dataset to a receiver for AI training, the representation capability of the trained model depends only on the distribution of the reconstructed data at the receiver. Then, our problem formulation is still applicable. Specifically, instead of requiring the exact reconstruction of the source data, it suffices for the perceptual constraint of our formulation to align the distribution of the system output $\hat{\boldsymbol{s}}$ with that of the original dataset $\boldsymbol{s}$. For this purpose, divergence measures such as the Kullback-Leibler (KL) divergence can be used as the perceptual metric $p(\cdot)$ to assess how closely the two distributions match. Besides, the task-relevant constraint is now used to ensure that the critical information required for the model training task is accurately preserved.



\section{Practical Implementation}\label{practical}
This section presents practical implementations of the AGI-driven GSC framework in two scenarios: online meetings and road monitoring, as depicted in Fig. \ref{semantic_practical}. For clarity, the data flows for the online meeting scenario are marked in red, whereas those for the road monitoring scenario are marked in black. The transmitter is placed on the left side of the “wireless transmission link”, and the receiver is placed on the right.


\subsection{Scenario 1: Online Meeting}
\begin{figure*}
  \centering
  \includegraphics[width=0.95\linewidth]{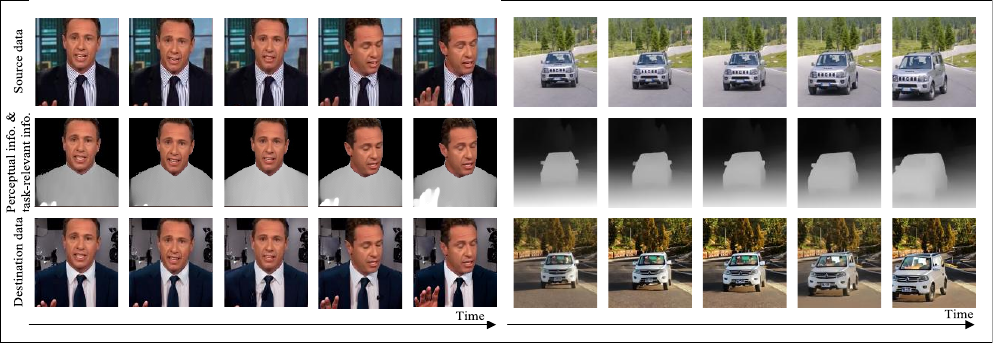}
  \caption{Data flow of the GSC framework under two scenarios. \textbf{Left Panel}: (Online meeting): Top row shows source data, middle row shows the composition of perceptual and task-relevant information, and bottom row shows destination data. \textbf{Right Panel}: (Road monitoring): Top row shows source data, middle row shows perceptual information, and bottom row shows destination data.}
  \label{semantic_experiment1}
\end{figure*}

We first consider the scenario of an online meeting. As shown in the upper half of Fig. \ref{semantic_practical}, a speech video is transmitted from the transmitter to the receiver, requiring the distortion in facial information of the video to be minimized. To achieve this, given the video as source data and the task objective of “sending face information and using depth maps as perceptual information”, an AI agent calls the segment-anything model in \cite{kirillov_segment_2023} to extract facial segments from the source data as task-relevant information. Non-facial segments are also extracted and then processed by the depth-anything model in \cite{yang_depth_2024} to produce depth maps as perceptual information. Subsequently, both types of information are compressed using principal component analysis (PCA), then encoded with 5G low-density parity check (LDPC) code, and transmitted over a wireless additive white Gaussian noise (AWGN) channel with a signal-to-noise ratio (SNR) of 10 dB. To reduce communication overhead, details such as textures and colors are excluded during transmission. At the receiver, after LDPC decoding and PCA decompression, the semantic decoder relies on a stable diffusion model \cite{chen_control--video_2023}, guided by a ControlNet \cite{zhang_adding_2023}, to generate the destination data that meets the task-relevant constraint and perceptual constraint in our problem formulation.

We perform simulations to validate the implementation of GSC. In the left panel of Fig. \ref{semantic_experiment1}, three rows, from top to bottom, successively display frame samples from the source data, the composition of both task-relevant and perceptual information, and the destination data. We see that the face segments in the destination data closely match the task-relevant information. This indicates that the distortion in task-relevant information is negligible, satisfying the task-relevant constraint. We also see that the destination data aligns well with human perception, meeting the perceptual constraint.


\subsection{Scenario 2: Road Monitoring}
We now consider the road monitoring scenario, in which a monitoring device captures a road video as the source data and transmits it to the receiver, as depicted in the lower half of Fig. \ref{semantic_practical}. The task objective in this scenario is to “describe the video and ensure the generated video matches the description.” The practical implementation of GSC in this scenario is similar to that in the online meeting scenario. Unlike the online meeting scenario, here the AI agent calls the BLIP2 model \cite{li_blip-2_2023} to generate a textual video description as task-relevant information, which is then encoded using UTF-8 instead of PCA. Similar to the online meeting scenario, we conduct simulations to validate the implementation of the GSC framework. As shown in the right panel of Fig. \ref{semantic_experiment1}, the three rows from top to bottom display video samples from the source data, the perceptual information, and the destination data. In the destination data, we observe a white car driving on the road, which is consistent with the video description and aligns well with human perception.


In the road monitoring scenario, we compare our AGI-driven GSC with the recent generative semantic communication scheme in \cite{grassucci2023generative} and a traditional communication baseline. The traditional communication scheme compresses source data with JPEG2000, applies 5G LDPC encoding, and sends it bit-by-bit. In \cite{grassucci2023generative}, semantic information is first extracted and then compressed with JPEG2000 and encoded with 5G LDPC for transmission. Table \ref{tab:compression-transposed} shows the performance of the three schemes using the semantic-NMSE, PIQE, and NRQM metrics defined in Section \ref{performance}. In terms of semantic-NMSE, the traditional communication scheme performs best, with error decreasing from 0.0521 to 0.0052 as the number of transmitted bytes increases. Our AGI-driven GSC follows, holding steady at about 0.0133 across all compression levels. The scheme in \cite{grassucci2023generative} shows a much higher semantic-NMSE of approximately 0.59. Additionally, the perceptual metrics PIQE and NRQM exhibit the opposite ordering. In terms of perceptual quality, the scheme in \cite{grassucci2023generative} achieves the best, with PIQE around 28 and NRQM about 19. Our AGI-driven GSC follows with PIQE near 35 and NRQM close to 19. The traditional communication scheme performs worst in perceptual quality, with PIQE around 55 and NRQM about 41. These results indicate that AGI-driven GSC strikes an attractive balance between task-relevant and perceptual metrics, making it well-suited for AGI applications.



\begin{table}[ht]
\centering
\renewcommand{\arraystretch}{1.3}
\setlength{\tabcolsep}{8pt}
\scriptsize
\caption{Performance Comparison in Terms of Semantic-NMSE, PIQE, and NRQM Metrics for Different Compression Rates}
\begin{tabular}{c|c|c|c|c}
\hline
\multirow{2}{*}{\textbf{Method}} & \multirow{2}{*}{\textbf{Metric}} & \multicolumn{3}{c}{\textbf{Total bytes transmitted ($\times 10^4$ bytes)}} \\
\cline{3-5}
 &  & \textbf{$\leq$ 7.86} & \textbf{$\leq$ 23.6} & \textbf{$\leq$ 39.3} \\
\hline
\multirow{3}{*}{\begin{tabular}{@{}c@{}}Traditional\\Comm.\end{tabular}}
 & Semantic-NMSE $\downarrow$ & 0.0521 & 0.0087 & 0.0052 \\
 & PIQE $\downarrow$ & 67.2336 & 52.5498 & 44.7033 \\
 & NRQM $\downarrow$ & 42.1322 & 40.1901 & 40.9679 \\
\hline
\multirow{3}{*}{GSC in \cite{grassucci2023generative}} 
 &Semantic-NMSE $\downarrow$ & 0.6109 & 0.5871 & 0.6143 \\
 & PIQE $\downarrow$ & 28.5834 & 27.7499 & 27.3332 \\
 & NRQM $\downarrow$ & 20.8965 & 21.7601 & 16.4710 \\
\hline
\multirow{3}{*}{\begin{tabular}{@{}c@{}}\textbf{AGI-driven}\\\textbf{GSC}\end{tabular}}
 &Semantic-NMSE $\downarrow$ & 0.0133 & 0.0133 & 0.0132 \\
 & PIQE $\downarrow$ & 37.0699 & 36.9293 & 33.5680 \\
 & NRQM $\downarrow$ & 18.8181 & 18.7836 & 18.7055 \\
\hline
\end{tabular}
\label{tab:compression-transposed}
\end{table}

We further evaluate the computational cost in floating-point operations (FLOPs) for each scheme. The traditional communication baseline, which uses JPEG2000 compression and 5G LDPC encoding, requires $3.63\times10^{10}$ FLOPs. The scheme in \cite{grassucci2023generative} follows the same coding methods, i.e., JPEG2000 and LDPC, and adds a diffusion model for decoding, requiring $6\times10^{12}$ FLOPs. Our AGI-driven GSC first uses BLIP-2 and the depth-anything model to extract semantic information and then applies a diffusion model at the receiver, resulting in $9.1\times10^{12}$ FLOPs. This means our method, similar to the scheme in \cite{grassucci2023generative}, needs about one hundred times more FLOPs than the traditional scheme.

\section{Conclusions and Open Challenges}\label{challenges}
In this article, we introduced the GSC framework, a new communication paradigm designed to support AGI-driven applications. We presented the key components of the GSC framework, including foundation models and generative models, and defined the fundamental problem of the GSC framework. We then elaborated on a practical implementation of GSC and illustrated its advantages for AGI applications through two preliminary use cases. Yet, since the development of GSC is still in its infancy, it is of utmost importance to tackle the open challenges that promote its further development.

The physical-layer constraints were intentionally omitted for simplicity in our discussions. Nevertheless, when implementing GSC in real-world scenarios, the physical-layer constraints become pivotal in the optimization of GSC. For example, the integration of multiple-input multiple-output (MIMO) techniques into the physical-layer air interface of GSC imposes transmission power constraints, along with additional optimization variables like beamforming vectors and precoding matrices. Similarly, the employment of the reconfigurable intelligent surface (RIS) technology in the air interface imposes additional constraints related to reflection elements. Moreover, considering the potential of non-orthogonal multiple access in the physical layer, the suppression of inter-user interference also needs to be considered. These factors significantly complicate the GSC system design, and meanwhile provide new research opportunities.

Moreover, GSC can be deployed in edge intelligence networks to enhance communication efficiency at the edge. In this scenario, a key challenge is that these edge devices typically have limited computational capacity, making it difficult to accommodate the foundational and generative models required by GSC. To tackle this issue, one approach is to develop lightweight large-scale models using techniques such as distillation, quantization, and pruning, at the cost of performance degradation. Alternatively, a more promising approach is to distribute the computational load of these large-scale models across the edge devices, thereby reducing the burden on an individual device. In this case, it is crucial to explore distributed realizations of these large-scale models. Additionally, edge intelligence-related constraints, such as hardware resource limitations of devices, transmission delay constraints, and device energy consumption constraints, need to be considered. These factors increase the difficulty of the GSC design and provide prosperous research opportunities.

We hope that the challenges and opportunities described above will help pave the way for researchers in the development of AGI-driven GSC in the future.

\bibliographystyle{IEEEtran}
\bibliography{ref}

\needspace{\baselineskip}

\section*{Biographies}

\vspace{-33pt}
\small
\begin{IEEEbiographynophoto}{Xiaojun Yuan}
(Senior Member, IEEE) received the Ph.D. degree in Electrical Engineering from the City University of Hong Kong in 2009. He is now a state-specially-recruited professor with the University of Electronic Science and Technology of China. His research interests cover a broad range of signal processing, machine learning, and wireless communications, including but not limited to intelligent communications, structured signal reconstruction, Bayesian approximate inference, distributed learning, etc. He was an editor of IEEE leading journals, including \textsc{IEEE Transactions on Wireless Communications} and \textsc{IEEE Transactions on Communications}. He was a co-recipient of the Best Paper Award of IEEE International Conference on Communications (ICC) 2014, a co-recipient of the Best Journal Paper Award of IEEE Technical Committee on Green Communications and Computing (TCGCC) 2017, and a co-recipient of IEEE Heinrich Hertz Award for Best Communication Letter 2022.
\end{IEEEbiographynophoto}

\vspace{-33pt}
\begin{IEEEbiographynophoto}{Haoming Ma}
received the B.Eng. degree in electronic information engineering from the School of Information and Communication Engineering, University of Electronic Science and Technology of China (UESTC), in 2021. He also received the M.S. degree in communication engineering from the National Key Laboratory of Wireless Communications, UESTC, in 2024. His research interests include turbo compressed sensing, distributed learning and semantic communication.
\end{IEEEbiographynophoto}

\vspace{-33pt}
\begin{IEEEbiographynophoto}{Yinuo Huang}
received the B.Eng. degree from the Yingcai Honors College, University of Electronic Science and Technology of China (UESTC), in 2023. He is currently pursuing the Ph.D. degree with the National Key Laboratory of Wireless Communications, UESTC. His research interests include wireless communications and machine learning. 
\end{IEEEbiographynophoto}

\vspace{-33pt}
\begin{IEEEbiographynophoto}{Zhoufan Hua}
received the B.Eng. degree in electronic information engineering from the School of Information Engineering, Zhengzhou University, in 2023. He is currently pursuing the M.S. degree with the National Key Laboratory of Wireless Communications, University of Electronic Science and Technology of China. His research interests include semantic communication.
\end{IEEEbiographynophoto}

\vspace{-33pt}
\begin{IEEEbiographynophoto}{Yong Zuo}
(Member, IEEE) received the B.E. degree in communication engineering from Zhejiang University, Hangzhou, China, and the Ph.D. degree in communication engineering from the Chinese Academy of Sciences, Beijing, China. From 2012 to 2015, he was a Research Scientist with ALU Bell Labs, where he worked on 5G standardization, network optimization algorithm design and AI fields. He is currently a Professor with Xiangjiang Laboratory, Changsha, China. He is also an Adjunct Professor with the National University of Defense Technology and Zhejiang Lab. His research interests include AI, satellite communications, internet of things and wireless communications, including but not limited to distributed learning, intelligent communications, and bayesian approximate inference.
\end{IEEEbiographynophoto}

\vspace{-33pt}
\begin{IEEEbiographynophoto}{Zhi Ding}
(Fellow, IEEE) is with the Department of Electrical and Computer Engineering at the University of California, Davis, where he holds the position of distinguished professor. His major research interests and expertise cover the areas of wireless networking, communications, signal processing, multimedia, and learning. Prof. Ding is a Fellow of IEEE and currently serves as the Chief Information Officer and Chief Marketing Officer of the IEEE Communications Society. He was also an IEEE \textit{Distinguished Lecturer} (Circuits and Systems Society, 2004-06, Communications Society, 2008-09). He served on as \textsc{IEEE Transactions on Wireless Communications} Steering Committee Member (2007-2009) and its Chair (2009-2010). Dr. Ding is a coauthor of the textbook: \textit{Modern Digital and Analog Communication Systems}, 5th edition, Oxford University Press, 2019. Prof. Ding received the IEEE Communication Society’s WTC Award in 2012 and the IEEE Communication Society’s Education Award in 2020.
\end{IEEEbiographynophoto}

\end{document}